\documentclass[conference,a4paper]{APSIPA2021}
\usepackage{amsmath}
\usepackage{amssymb}
\usepackage{multirow}
\usepackage{threeparttable}

\usepackage{afterpage}
\usepackage[pdftex]{graphicx}

\usepackage{geometry}
\usepackage{fancyhdr}

\fancypagestyle{firststyle}{
  \fancyhf{}
  \fancyhead[C]{2024 Asia Pacific Signal and Information Processing Association Annual Summit and Conference (APSIPA ASC)}
}

\newcommand\secref[1]{Section~\ref{#1}}
\newcommand\figref[1]{Fig. ~\ref{#1}}
\newcommand\tabref[1]{Table~\ref{#1}}
\newcommand\fmlref[1]{Equation~(\ref{#1})}

\begin{document}

\title{Enhancing Remote Adversarial Patch Attacks on Face Detectors with Tiling and Scaling}

\author{
\authorblockN{
Masora Okano\authorrefmark{1},
Koichi Ito\authorrefmark{2},
Masakatsu Nishigaki\authorrefmark{1} and
Tetsushi Ohki\authorrefmark{1,3}
}

\authorblockA{
\authorrefmark{1}
Shizuoka University, Shizuoka, Japan \\
E-mail: okano@sec.inf.shizoka.ac.jp, \{nisigaki, ohki\}@inf.shizuoka.ac.jp}

\authorblockA{
\authorrefmark{2}
Tohoku University, Miyagi, Japan \\
E-mail: ito@aoki.ecei.tohoku.ac.jp}

\authorblockA{
\authorrefmark{3}
RIKEN AIP, Tokyo, Japan}
}

\maketitle
\thispagestyle{firststyle}
\pagestyle{fancy}

\begin{abstract}
This paper discusses the attack feasibility of Remote Adversarial Patch (RAP) targeting face detectors.
The RAP that targets face detectors is similar to the RAP that targets general object detectors, but the former has multiple issues in the attack process the latter does not.
(1) It is possible to detect objects of various scales. In particular, the area of small objects that are convolved during feature extraction by CNN is small,so the area that affects the inference results is also small.
(2) It is a two-class classification, so there is a large gap in characteristics between the classes. This makes it difficult to attack the inference results by directing them to a different class.
In this paper, we propose a new patch placement method and loss function for each problem. The patches targeting the proposed face detector showed superior detection obstruct effects compared to the patches targeting the general object detector.
\end{abstract}

\section{Introduction}
\label{sec:introduction}

Deep neural network (DNN) models are susceptible to malicious manipulation of input data. Previous researches have employed a variety of approaches, 
including the use of specially designed spectacles that obscure key feature points of an object by reflecting light \cite{Yamada2013-uo}, as well as the development of sophisticated Adversarial Examples (AE) that subtly alter pixel values to cause the model to misclassify inputs intentionally. 
The threat of DNN attacks has emerged as a significant concern in light of the increasing incorporation of DNN-based models in a multitude of systems. 
Additionally, researches have explored the Remote Adversarial Patch (RAP), an attack that does not depend on the specific object being targeted. Unlike methods that directly alter features, RAP can remotely influence the model's output.

RAP was defined by Mirsky et al\cite{mirsky2023ipatch}. They identified two key characterisitics in their definition of RAP.
One characteristic is the semantic alteration of model inference caused by patches, and the other is the ability of patches to be remotely attacked.
Among these characteristics, the remote attackability of RAP is particularly significant.
Here, the word “remote” means that the AP exists outside the target object's area. Compared to an AP that is placed on the target object, a RAP that does not directly modify the target object's characteristics is more difficult from an attack perspective.
For the sake of simplicity, this paper defines RAP solely based on the characteristic of remote attackability.


This study focuses on the application of RAP to face detection. Although RAP has been extensively discussed for general object detection (multi-class object detection)\cite{Liu2018-vy,Lee2019-ab}, the feasibility of RAP for face detection remains underexplored. Focusing on face detection is particularly important because obstructing it significantly enhances privacy protection. For example, applying a RAP to images before publication or capture could prevent unauthorized third parties from replicating the face region.

Two key challenges in applying RAP to face detection are as follows.
\textbf{(1) the scale of detectable objects is diverse.}
In the field of face detection, unlike the general object detection field, there is a requirement to detect obscure and distant objects, such as surveillance camera images. As a result, there is a tendency to focus on small features so that even extremely small faces can be detected. Small features have limited characteristics in the area around the face in the image that are included in the convolution. Consequently, the area that affects face detection is also small. Thus, the restrictions on where the patches can be placed are strict and are considered difficult to attack by RAP.
\textbf{(2) fewer classes to classify.}
Face detection involves fewer classes compared to multi-class object detection. In addition, because RAP operates from a distance and does not alter the object's intrinsic features directly. Therefore, it is difficult to redirect a class given to a particular object to another class with similar features, which limits the approaches that can be taken in an attack.

To address this issue, this study proposes a novel RAP method which involves two main processes: scaling and tiling. The \textbf{scaling process} adjusts the size of patches to correspond with different face scales during training, enhancing optimization across varying scales. 
The \textbf{tiling process} arranges the patches in a grid pattern, ensuring that any cropped region of the image will contain part of a patch. 
This approach addresses the issue by applying a tiling process that ensures any cropped region contains patches to some extent, and a scaling process that adjusts their relative size according to the scale variation of the face.

In addition to optimizing patches using the proposed patch applying method, we introduce a novel loss function for face detection obfuscation, called the \textbf{Borderline False Positives Loss}. The Borderline False Positives Loss is designed to increase the number of false positives near the true face region and to disturb the coordinates of the true face. This enables false inferences on the face detection model without forcing the inference class towards the background class, which is significantly different from the face class. In other words, we overcome the attack limitations of having (2) fewer classes to classify, thereby realizing an effective attack method.

We conducted comprehensive experiments on the proposed method using multiple datasets and various RAP methods. Through these experiments, we demonstrated the effectiveness of the proposed method in terms of its obstruction performance against patch-based face detectors and its performance across diverse scales.

The contributions of this paper are listed below.
\begin{itemize}
    \item We proposed a novel RAP method based on patch tiling and borderline false positive loss.
    \item Our comprehensive experiments showed consistent obstruction performance.
    \item Our proposed method also demonstrated consistent obstruction performance across datasets with varying face scales, showing robustness to face scale variation.
\end{itemize}


\begin{figure*}[ht]
\centering
\includegraphics[width=0.95\textwidth]{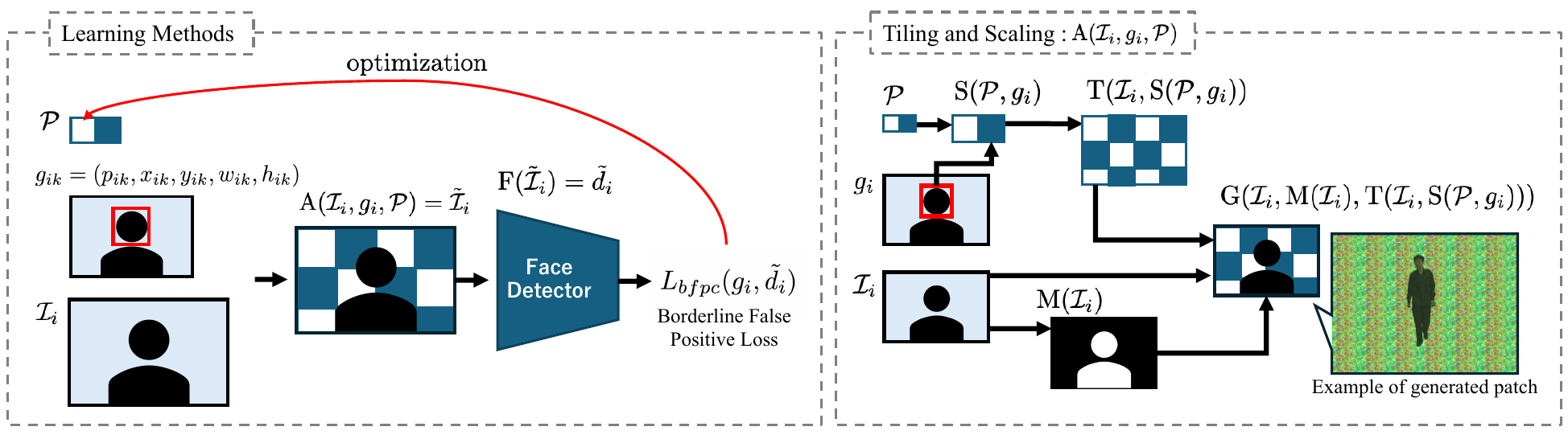}
\caption{Schematic diagram of the proposed patch generation method. We propose a patch application method and its learning method to explore the feasibility of generating RAPs to attack face detectors}
\label{fig:structure}
\end{figure*}

\section{Related work}

\subsection{Adversarial Attack}
Data that exploits vulnerabilities in a Deep Neural Network (DNN) model when attached to an image is called an Adversarial Patch (AP). Although AP generation methods were initially discussed for image classification models, Song et al. proposed the first AP generation method for object detection models \cite{song2018physical}. Compared to image classification tasks, object detection tasks detect and label multiple objects in a scene, making it more difficult for AP to obstruct detection. APs have also been studied for person detection tasks\cite{Thys2019-fo}. Person detection is considered to be difficult to obstruct due to the large diversity within the person class.

However, all the studies discussed so far had the limitation that they must be located on a defined region of the detected object.To overcome the region limitation, Liu et al. proposed an adversarial patch, called DPatch\cite{Liu2018-vy}, so that the effect of the attack is independent of the location.Their method is similar to that of Liu et al. but differs in that Liu et al. optimize the patch so that its region of presence is the only region proposed, whereas Lee et al. maximize the losses used by the model during training\cite{Lee2019-ab}.

On the other hand, Mirsky et al. proposed a patch generation method that can apply AP to segmentation models and simultaneously defined RAP\cite{mirsky2023ipatch}. According to the definition given by Mirsky et al., the methods of Lee et al. and Liu et al. do not strictly fit the definition of RAP. However, in this paper, we define RAP as a method that can be attacked remotely, and therefore include these two methods as RAP.

\subsection{Face Detection Obstruction}
Obstruction of face detection has been discussed for the purpose of preventing unauthorized face image leakage due to unintentional capture of face images. Yamada et al. \cite{Yamada2013-uo} proposed a detection jamming method that does not interfere with facial expression communication, in which facial features are modified by glasses irradiating near-infrared signals. AE \cite{Zhang2022-ak} and AP \cite{Yang2020-si} in face detection tasks have also been studied, but all of these methods involve processing of the face areas to be disturbed. Therefore, their applications are limited for privacy protection purposes.
On the other hand, the patches generated by the proposed method do not require processing of face regions.


\section{Method}

The proposed method consists of a patch application process and a learning process. Each process is explained in the sections \secref{subsec:learning_methods} and \secref{subsec:patch_apply}, respectively. The overall process is shown in \figref{fig:structure}.
Let $I$ be a dataset consisting of $N$ images, where each image can be represented as $\mathcal{I}=\{I_i \mid i=0,\cdots,N-1\}$. The face detector $F$ takes an image $I_i$ as input and returns an inference result $d_i=\{d_{ik} \mid k=0,\cdots,M-1\}$. 
Here, the inference result $d_i$ consists of $M \geq 0$ face regions $d_{ik}$. Each face region $d_{ik}$ includes the rectangular coordinates $(x, y)$ of the rectangle's center, the width and height $(w, h)$, as well as the confidence value $p$ for the detection. It can be expressed as follows:
\begin{eqnarray}
d_{ik}=(p_{ik},x_{ik},y_{ik},w_{ik},h_{ik}).
\end{eqnarray}
In addition, the Ground Truth (GT) of the correct face detection inference result for each image is $g_i$.

Let $P$ be a adversarial patch, and let $(w_{P},h_{P})$ be its width and height. Using the embedding function $A$ to embed the patch in the image, the patched image $I_i$ can be expressed as follows.
\begin{eqnarray}
\Tilde{I}_i = A(I_i,P).
\end{eqnarray}
%
Furtermore, let the face detection results for $\Tilde{I}_i$ be denoted as $\Tilde{d}_i$, with each element expressed similarly to $d_i$ as follows.
\begin{eqnarray}
    \Tilde{d}_i &=& \{\Tilde{d}_{ij} \mid j=0,\cdots,M-1\} \\
    \Tilde{d}_{ij} &=&(\Tilde{p}_{ij},\Tilde{x}_{ij},\Tilde{y}_{ij},\Tilde{w}_{ij},\Tilde{h}_{ij}).
\end{eqnarray}


\subsection{Learning Methods}
\label{subsec:learning_methods}

\subsubsection{Definition of Obstruction}
For the detection $\Tilde{d}_{ij}$, True Positive (TP) and False Positive (FP) are defined based on the IoU threshold value $\theta_D$, and the number of cases where the detection of the ground truth (GT) fails is defined as False Negative (FN), as follows.
\begin{eqnarray}
\label{fml:tp}
\Tilde{d}_{ij} \text{ is TP} &\iff& \exists g_{ik} \text{ such that } IoU(\Tilde{d}_{ij}, g_{ik}) \geq \theta_D \\
\label{fml:fp}
\Tilde{d}_{ij} \text{ is FP} &\iff& \forall g_{ik} \text{ such that } IoU(\Tilde{d}_{ij}, g_{ik}) < \theta_D \\\label{fml:fn}
g_{ik} \text{ is FN} &\iff& \forall \Tilde{d}_{ij} \text{ such that } IoU(\Tilde{d}_{ij}, g_{ik}) < \theta_D
\end{eqnarray}

True Negative (TN) indicates whether the system can correctly identify areas where faces do not exist. However, in this paper, we will be focusing on the True Positive (TP), False Positive (FP) and False Negative (FN) values related to the face areas detected by the detector. 
A decrease in TP indicates that the face detector is failing to correctly identify face areas, while an increase in FP suggests that it is becoming more difficult to extract the correct face areas from the detection results. 
Thus, the efficiency of obstructing face detection can be evaluated based on the decrease in TP and the increase in FP.

\subsubsection{Borderline False Positive Loss}

%

In this study, we propose \textbf{Borderline False Positive Loss} to reduce True Positives (TP) and increase False Positives (FP). This loss function increases FPs around the boundary of the rectangle in $g_i$ by obstructing the true face area, thereby reducing TP.
%
Let $\mathcal{L}_{bfpc}$ denote the Borderline False Positive Loss, \fmlref{fml:objective_function}, the objective function, \fmlref{fml:bfp_loss} the definition of loss function.
\begin{gather}
\label{fml:objective_function}
\min_{P} \{\mathcal{L}_{bfp}(g_i,\Tilde{d}_i)\}. \\
\label{fml:bfp_loss}
\mathcal{L}_{bfpc}(g_i,\Tilde{d}_i) = -\sum^{M}_{j=0} b_{ij} * log(1-\Tilde{p}_{ij}).
\end{gather}
Here, $b_{ij}$ is shown in the formula for the borderline variable in the \fmlref{fml:borderline_variable}. $b_{ij}$ is a borderline judgment variable that is 1 when the inference result for the image with the patch added, $\Tilde{d}_{ij}$, is $\theta_{T} > a_{ij} \geq \theta_{F}$, which is the boundary between TP and FP, and 0 otherwise.
\begin{equation}
b_{ij} =
    \begin{cases}
        0 & (a_{ij} \geq \theta_{T})~or~(a_{ij} < \theta_{F}) \\
        1 & \theta_{T} > a_{ij} \geq \theta_{F}
    \end{cases}.
\label{fml:borderline_variable}
\end{equation}
The values of $\theta_{T}$ and $\theta_{F}$ are IoU thresholds, set separately from the threshold $\theta_D$ used to classify $\Tilde{d}{ij}$ as TP or FP. 
The following inequality must hold for each threshold: $\theta_{T}>\theta_D>\theta_{F}$.
As shown in \fmlref{fml:iou_max}, $a_{ij}$ represents the maximum IoU value calculated between $g_i = \{g_{ik} \mid k=0, \cdots, M-1\}$ and $\Tilde{d}_{ij}$.
\begin{align}
a_{ij} = \max_{g_{ik} \in g_i} IoU(g_{ik}, \Tilde{d}_{ij}).
\label{fml:iou_max}
\end{align}

The loss function, $\mathcal{L}_{bfpc}$, increases the confidence value of the inference results that may be false positives based on the borderline decision variable, $b_{ij}$.
The boundary judgment variable $b_{ij}$ is a variable that classifies the inference results to be the judgment boundaries between TP and FP based on the maximum IoU value $a_{ij}$ obtained between $g_i$ and $\Tilde{d}_{ij}$.
The loss function, $\mathcal{L}_{bfpc}$, is designed to mislead the coordinates of the inference results that would otherwise be true positives into false positives by increasing the number of false positives around the true face region using $b_{ij}$.

\subsection{Patch Applying Process}
\label{subsec:patch_apply}






The expression for the patch application process $A$ is shown as \fmlref{fml:patch_apply}, and the process of the patch application process is shown in the figure \figref{fig:structure} on the right.
\begin{align}
A(I_i, g_i, P)=G(I_i, M(I_i), T(I_i, S(P,g_i))).
\label{fml:patch_apply}
\end{align}
%
The function $M$ takes the image $I_i$ as an argument and converts it into a reference image for masking only the foreground.
The function $G$ takes as arguments three images(foreground, mask and background) of equal height and width, and performs foreground-background composition.
The scaling function $S$ and the tiling function $T$ are both functions that perform transformations on the patches that are optimized.
In applying patches, the scaling and tiling processes play a particularly important role.

\subsubsection{Scaling}
The scaling process is the $S(P, g_i)$ on the rigth side of \figref{fig:structure}
. By making the scaling function $S$ align the area ratio of patches and faces, it encourages patches to be able to obstruct faces of various scales uniformly during learning.
In the scaling process, patch $P$ is resized to have a width of $w_{P^S}$ and a height of $h_{P^S}$.
Using patch $P$ and the corresponding GT $g_i$, the width $w_{P^S}$ and height $h_{P^S}$ of the scaled patch are expressed by the following equations.
%
%
\begin{eqnarray}
    (w_{P^S}, h_{P^S}) &=& S(P, g_i)  \nonumber \\
    &=& (round(w_{P} * s), round(h_{P} * s)). \\
    s &=& \sqrt{\frac{\alpha * w_{ik} * h_{ik}}{h_P * w_P}}. \\
    k &=& \arg\max_{k} w_{ik} * h_{ik}.
\end{eqnarray}


\subsubsection{Tiling}
The $T(I_i, S(P, g_i))$ on the right side of \figref{fig:structure} is a tiling process. The tiling function $T$ is a process that converts a patch $P^{S}$ into a patch tile $P^{T}$ of the same height and width as the image by tiling the patch $P^{S}$ horizontally and vertically. 
Reducing the dependence of face detection on face position by ensuring that the pixels of the patch are included when any face in any area is extracted as a feature.
Each coordinate $P^T[x, y]$ of $P^T$ can be determined using the tiling function $T$ as follows.
\begin{eqnarray}
    P^T &=& T(I, P^S) .\\
    P^T[x, y] &=& P^S[x \bmod w_{P^S} , y \bmod h_{P^S}] .
\end{eqnarray}
%
The edges of the patches created by tiling will be cut off to match the image size.


\section{Experiment}

\subsection{Experimental Setup}
\label{sec:experimental_setup}

\subsubsection{Dataset}
Two datasets are utilized in the experiment: CASIA Gait B (CGB) \cite{Zheng2011-vp} and FaceForensics++ (FFP) \cite{Rossler2019-jr}. CGB is designed for gait recognition and includes images captured in a controlled indoor environment.
The dataset features 124 subjects, each captured from 11 different angles. FFP, on the other hand, is a dataset for detecting DeepFake videos and comprises 1000 videos featuring frontal faces without occlusion, collected from Youtube press conference videos. 
In the experiment, only frontal faces were extracted from the CGB to match the experimental conditions with the FFP containing frontal faces. Images without facial regions were removed from these videos using S3FD to avoid duplication, and the training and validation datasets of 3000 images each were extracted.
In addition, for the purpose of detection obstruction, GT is the inference result $d_i$ of the image $I_i$ before patch application. Therefore, in the experiment, $d_i=g_i$.

\subsubsection{Preprocessing}
For these moving images, a mask image and $\hat{D}_i$, the GT, are created in advance. In this experiment, rembg\cite{Gatis_undated-wl} was used to create the mask image and S3FD was used to create $\hat{D}_i$.

\subsubsection{Learning Patch}
For the various hyper-parameters, the IoU threshold $\theta_D=0.5$, and the scaling parameter $\alpha=5.58$ and the loss function parameters $\theta_T=0.6,\theta_F=0.3$ are determined from empirical results. For optimization, the Nesterov Iterative Fast Gradient Sign Method (NI-FGSM)\cite{Lin2019-se} is used in the proposed method. In addition, the methods of DPatch\cite{Liu2018-vy} and Lee et al.\cite{Lee2019-ab} are used for comparison.

\subsubsection{Evaluation Methods}
The F value and the Average Precision (AP) value are used to compare how much the detection performance is degraded compared to the situation without patch obstructions. The number of TPs, FPs, and GTs are also used to compare obstruction performance. The reason for not simply using F and AP values is that F values are insensitive to changes in the number of TPs when the number of FPs becomes extremely large relative to the number of TPs, and AP values are insensitive to increases in the number of FPs, making it difficult to correctly compare obstruction performance.

\subsection{Detection Obstruction Performance Evaluation}
\label{sec:obstruction_performance}
We present a comparison of the proposed RAP generation method with other methods. DPatch\cite{Liu2018-vy} and Lee et al.'s method\cite{Lee2019-ab} are used for the comparison. The coordinates on the image where the two patches to be compared are placed are determined randomly, as in the case of both patches generation.S3FD is used for face detection. The experimental results are shown in \ref{tab:study_by_face_scale}.
\begin{table}[h]
 \caption{Comparison of detection obstruction performance}
 \centering
  \begin{tabular}{ll|lllll}
   \hline
   dataset & method & F & AP & GT & TP & FP \\
   \hline \hline
    \multirow{2}{*}{CGB} & Dpatch & 9.96e-1 & 1.0 & 3000 & 2977 & 0 \\
    & Lee & 1.34e-1 & 4.14e-2 & 3000 & 2967 & 38235 \\
    & Proposed & 7.91e-1 & 9.99e-1 & 3000 & \textbf{1967} & 7 \\ \hline
    \multirow{2}{*}{FFP} & Dpatch & 9.13e-1 & 9.87e-1 & 3315 & 3305 & 617 \\
    & Lee & 1.96e-1 & 9.29e-2 & 3315 & 3287 & 26866 \\
    & Proposed & 9.78e-1 & 9.99e-1 & 3315 & \textbf{3227} & 56 \\
   \hline
  \end{tabular}
\label{tab:study_by_face_scale}
\end{table}
%
\tabref{tab:study_by_face_scale} shows that the proposed method has fewer TP than other methods on both datasets. This indicates that Borderline False Positive Loss can successfully disturb the coordinate of the true face. However, as shown in \tabref{tab:study_by_face_scale}, the increase in FP with the proposed method is not as pronounced as in the method by Lee et al. This suggests that further increasing FP may enhance the effect of the Borderline False Positive Loss.



When CGB is used as the training data, the patches generated by the proposed method reduce TP when verified using CGB. However, when FFP is used to learn and verify patches, the degree of suppression of TP is less than when CGB is used. 
We think that the reason for this difference in the dataset is that CGB contains many small faces, while FFP contains almost large faces. For smaller faces included in the CGB, the IoU fluctuates steeply in overlapping regions, making it easier to meet the threshold requirements.



\subsection{Positional Robustness Evaluation}
\label{sec:positional_evaluation}

When detection is obstructed for different images, the relative positional fluctuations between the patch and the obstructed object due to the difference in the coordinates of the obstructed object for each image are considered to affect the detection obstruction performance. Therefore, we verify whether detection obstruction is robust to such position variations by comparing its position independence with the DPatch\cite{Liu2018-vy} and Lee et al.'s method\cite{Lee2019-ab}.
The inference results for all images in the test set with the patch applied are classified into TP and FP based on the ground truth, and the frequency for each upper right coordinate is tabulated and plotted as a heat map. The two patches to be compared are fixed at the coordinate $(0,0)$ to illustrate the effect of the patch on detection. Ideally, the verification should be performed on a dataset where the coordinates of the obstacle targets are evenly distributed throughout the image. In reality, however, it is difficult to prepare such a dataset. Therefore, we created a coordinate uniform data set that met the ideal conditions in a pseudo-way, and conducted a verification.

\begin{figure}[t]
\centering
\includegraphics[width=0.45\textwidth]{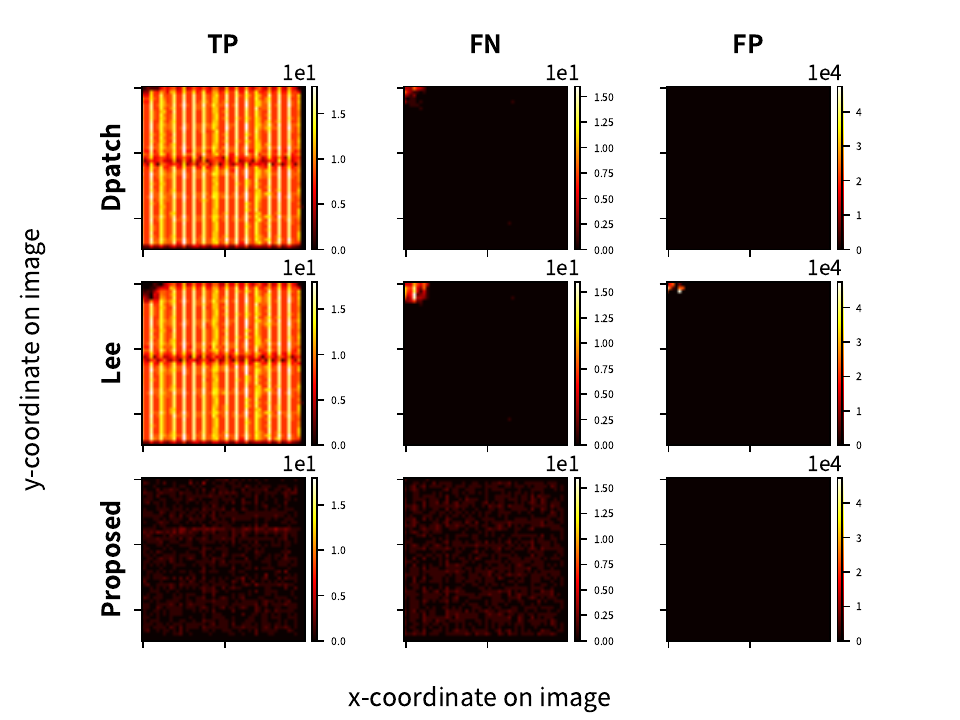}
\caption{Detection results for the coordinate uniform dataset. Top row: DPatch, middle row: patches by Lee et al. and bottom row: patches by the proposed method. From left to right: TP, FN, FP}
\label{fig:moving_det}
\end{figure}

\textbf{Coordinate Uniform Data Set} In order to show that it is possible to obstruct detection no matter where the face is located in the image, we created a dataset in which the positions of the face regions are uniformly distributed throughout the image, based on 10 images randomly extracted from the CASIA Gait B test set. The coordinate uniform dataset was created by shifting the image based on the top left coordinate of the face region and repeating the operation from the top left to the bottom right of the image. In this case, the width of each movement was set to 25 pixels. The coordinate-uniform dataset was created by inferring the face region before applying the tiling adversarial background image, and then excluding images for which there were no inference results. The total number of images in the dataset is 24,654.
The results of the experiment are shown in \figref{fig:moving_det}.
The verification results show that the Dpatch and Lee et al. patches each have a concentration of FNsin the vicinity of the coordinates where the patches are located $(x,y)=(0,0)$. In contrast, the proposed method distributes the FNs throughout the image. This result demonstrates the effectiveness of the proposed method's tiling process, and shows one advantage over previous research. This result is also one basis for supporting the hypothesis proposed in \secref{sec:introduction} that the placement coordinates of the patches are highly constrained when disturbing face detection.

\subsection{Dataset Transferability}
\label{sec:dataset_transferability}

For the purpose of protecting privacy, it is desirable that the generated patches can obstruct face detection in various situations. Therefore, we will swap the data sets between the training and validation phases, and verify whether the obstructing performance of the patches learned with the known data set can obstruct the unknown evaluation data set as well as the known training data set. If it is possible to obstruct face detection in various situations, it should be possible to obstruct other datasets with different shooting conditions in the same way as the training dataset. In addition, we will consider the possibility that the transferability to unknown datasets may differ depending on the face detection model used during training, and we will conduct verification using multiple models.

The patch is tested using a different dataset from the one used for learning.
We conduct experiments for each of the three models MTCNN\cite{Zhang2016-lr}, S3FD\cite{Zhang2017-sf}, and RetinaFace\cite{Deng2020-je} and compare them. The experimental results are shown in Table\ref{tab:dataset_transferability}.

\begin{table}[h]
 \caption{Table for dataset transferability}
 \centering
  \begin{tabular}{ll|lllll}
   \hline
   \shortstack{train/test \\ dataset} & model & F & AP & GT & TP & FP \\
   \hline \hline
    \multirow{2}{*}{CGB / CGB} & MTCNN & 2.92e-1 & 1.0 & 3000 & 513 & 0 \\
    & S3FD & 7.91e-1 & 9.99e-1 & 3000 & 1967 & 7 \\
    & RetinaFace & 7.87e-1 & 1.0 & 3000 & 1947 & 0 \\ \hline
    \multirow{2}{*}{CGB / FFP} & MTCNN & 9.14e-1 & 9.99e-1 & 3315 & 3135 & 407 \\
    & S3FD & 9.78e-1 & 9.99e-1 & 3315 & 3226 & 58 \\
    & RetinaFace & 9.72e-1 & 9.99e-1 & 3315 & 3147 & 10 \\ \hline
    \multirow{2}{*}{FFP / FFP} & MTCNN & 9.18e-1 & 9.99e-1 & 3315 & 3132 & 371 \\
    & S3FD & 9.78e-1 & 9.99e-1 & 3315 & 3227 & 56 \\
    & RetinaFace & 9.72e-1 & 9.99e-1 & 3315 & 3147 & 10 \\ \hline
    \multirow{2}{*}{FFP / CGB} & MTCNN & 2.92e-1 & 1.0 & 3000 & 512 & 0 \\
    & S3FD & 7.97e-1 & 9.99e-1 & 3000 & 1993 & 10 \\
    & RetinaFace & 7.92e-1 & 9.99e-1 & 3000 & 1968 & 1 \\
   \hline
  \end{tabular}
\label{tab:dataset_transferability}
\end{table}

The patches generated using CGB have kept the TP below 2,000 in the verification using CGB. In addition, patches generated using FFP kept the TP to about 3,200 in the verification using FFP.
All the models tended to have the same tendency in terms of transferability to unknown data sets. When referring to the results of verifying the patches learned with FP using CGB, TP is less than 2000 for all models. In contrast, when the same verification process is conducted using CGB with FFP, the results do not exhibit a similar trend.

The reason for this difference is thought to be that FFP contains many high-resolution images of faces of various sizes compared to CGB, and is more difficult as a detection obstruction task.
The patches trained on FFP have high-resolution and obvious facial features, which may have contributed to the enhancement of universal optimization independent of the dataset, although the optimization did not work well for the faces included in FFP.
As a result, it may have shown higher interference only for CGB containing low-resolution, ambiguous facial features.
Since the purpose of scaling is to find patterns that can obstruct the detection of faces of any scale, we believe that the obstructing performance of FFP against CGB is a result that shows the effectiveness of the scaling of the proposed method.


\section{Discussion}

\noindent \textbf{Multiple faces in the same scene :} 
In this experiment, we did not consider the case where there are multiple faces in the scene to be disturbed, but only the case where there is a single face.
This is because scenes where there is a single face are assumed to be online meetings and press conferences. By performing face detection obstruction on these videos, it is possible to protect the privacy of the photographed person, and it is effective.
We plan to consider the case where there are multiple faces in the future.

\noindent \textbf{Improvement of the loss function :} 
One possible reason for the smaller number of FP cases in \secref{sec:obstruction_performance} than in other methods is that the optimization process did not progress because the loss value often took the value 0 in the early stages of optimization.
In order for the value calculated by the Borderline False Positive Loss to be non-zero, the IoU calculated between the correct face region and the inferred detection needs to be less than $\theta_T=0.6$. On the other hand, for images without patches applied, the IoU will be close to 1.
Therefore, it is likely that the loss value did not change from 0 and optimization did not progress. To solve this, it seems that it is necessary to have an objective function that reduces the iou between the correct face region and the estimated detection results from 1 to around 0.6.
\section{Conclusion}

We discussed the difficulty of implementing RAP that targets face detectors and presented the diversity of face region scales and the small number of classification classes as reasons for this. In contrast, this study proposed a unique patch placement method and loss function as a solution to each of these problems. 

As a result of patching using the proposed method, it showed fewer TP than other methods. 
In addition, the patches learned on a dataset that tends to include large faces showed a certain level of obstruction performance on a dataset that tends to include small faces, indicating that they can cope with the diversity of face sizes.
Comparison with patches generated by other methods shows that the proposed disturbance method contributes to TP reduction even for face detectors with a small number of classification classes.

The proposed method still has room for improvement in both TP and FP, and further examination is needed regarding the refinement of the loss function and parameter optimization.

\section*{Acknowledgement}
This work was supported in part by JSPS KAKENHI JP 23H00463, JP 23K28084 and JST Moonshot R\&D Grant Number JPMJMS2215.

\bibliographystyle{ieeetr}
\bibliography{mybib}
\end{document}